\documentclass[letterpaper, 10 pt, journal, twoside]{IEEEtran}

\IEEEoverridecommandlockouts                              

\usepackage{graphicx} 
\usepackage{epsfig} 
\usepackage{mathptmx} 
\usepackage{times} 
\usepackage{amsmath} 
\usepackage{amssymb}  
\usepackage{tabularx}
\title{
  Omnidirectional DSO: 
  \\
  Direct Sparse Odometry with Fisheye Cameras
}

\author{Hidenobu Matsuki$^{1,2}$,  Lukas von Stumberg$^{2,3}$, Vladyslav Usenko$^{2}$, J\"org St\"uckler$^{2}$ and Daniel Cremers$^{2}$%
\thanks{Manuscript received: February, 23, 2018; Revised April, 16, 2018; Accepted June, 12, 2018.}
\thanks{This paper was recommended for publication by Editor Cyrill Stachniss upon evaluation of the Associate Editor and Reviewers' comments. }
\thanks{$^{1}$The University of Tokyo, Japan 　
{\tt\footnotesize  matsuki.hidenobu@gmail.com}}
\thanks{$^{2}$Technical University of Munich, Germany 
{\tt\footnotesize \{matsuki,stumberg, usenko, stueckle, cremers\}@in.tum.de}}
\thanks{$^{3}$Artisense GmbH, Germany}
\thanks{Digital Object Identifier (DOI): see top of this page.}
}

\begin{document}

\maketitle

\begin{abstract}

We propose a novel real-time direct monocular visual odometry for omnidirectional cameras.
Our method extends direct sparse odometry (DSO) by using the unified omnidirectional model as a projection function, which can be applied to fisheye cameras with a field-of-view (FoV) well above 180 degrees.
This formulation allows for using the full area of the input image even with strong distortion, while most existing visual odometry methods can only use a rectified and cropped part of it.
Model parameters within an active keyframe window are jointly optimized, including the intrinsic/extrinsic camera parameters, 3D position of points, and affine brightness parameters.
Thanks to the wide FoV, image overlap between frames becomes bigger and points are more spatially distributed.
Our results demonstrate that our method provides increased accuracy and robustness over state-of-the-art visual odometry algorithms.

\end{abstract}


\section{INTRODUCTION}
\IEEEPARstart{V}{isual} odometry (VO) with monocular cameras has widespread applications, for instance, in autonomous driving, mobile robot navigation or virtual/augmented reality. 
The benefit of using only a monocular vision system is that only a single low-cost camera is needed which is simple to maintain and often available in commodity hardware. Hence, research in the field of VO with monocular cameras is actively pursued in recent years~\cite{nister2004visual,forster2014svo,Semidense}.
Since VO algorithms estimate 3D structure and 6-DoF camera motion from visual information, sufficient texture needs to be present in the images so that correspondences can be observed between different frames.
A major limiting factor for correspondence estimation is the  field-of-view (FoV) of the camera.
This becomes especially apparent in environments with sparse features such as indoor environments with textureless walls, or dynamic environments where robust tracking requires that the static part of the environment is sufficiently visible in the frames.
Thus, wide FoV cameras are beneficial for VO.

It is, however, not straightforward to make full use of the wide FoV images in standard VO pipelines.
Typically, these approaches are designed for the pinhole camera model.
It projects measurements of 3D points onto an image plane and causes strong distortions of the image for a FoV of more than approx. 150 degrees.
To avoid the processing of distorted image regions, the images are typically cropped to a smaller part in the inner region, resulting in an effectively lower FoV.

There are two main approaches to increase the FoV of a VO system:
Firstly, optimization can be performed in a window of frames in which frames share varying mutual image overlap.
Examples are fixed-lag smoothing approaches such as the multi-state constrained Kalman Filter (MSCKF~\cite{mourikis2007_msckf}) or Direct Sparse Odometry (DSO~\cite{DSO}).
For some approaches, the use of a camera projection model such as the unified omnidirectional model can be a viable option to avoid image distortions.
In this paper, we propose a state-of-the-art direct visual odometry method that incorporates the unified omnidirectional model as used in~\cite{Omni} into a fixed-lag smoothing approach to VO.

\begin{figure}[tb]
     \begin{center}
		\includegraphics[width=8.5cm]{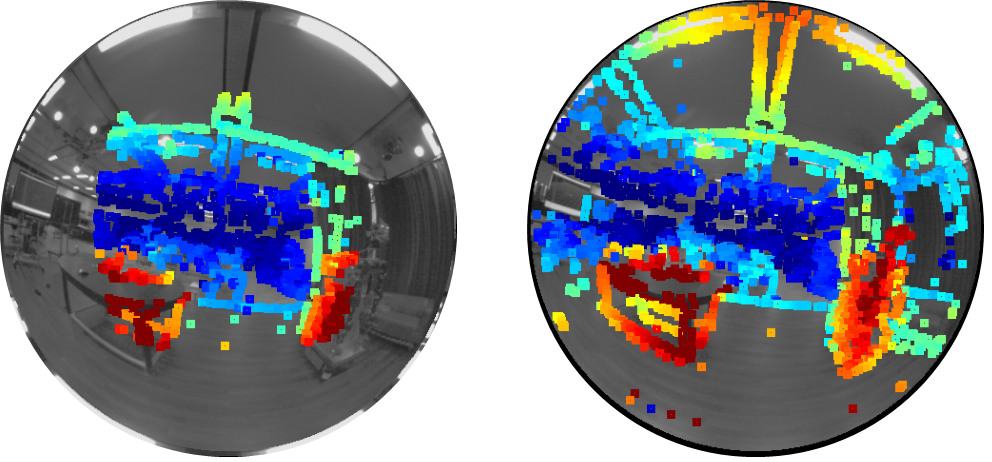}
				\caption{{\bf{Top: }} Active pixels in DSO for different camera models (left: pinhole, right: omnidirectional). With the pinhole model, DSO can only use a small part of the original fisheye image. In contrast, with the omnidirectional model, DSO can use pixels across the whole fisheye image.
				{\bf{Bottom: }}Example of reconstructed map
				}
		\label{top}
		        \end{center}
        \begin{center}
		\includegraphics[width=7.0cm]{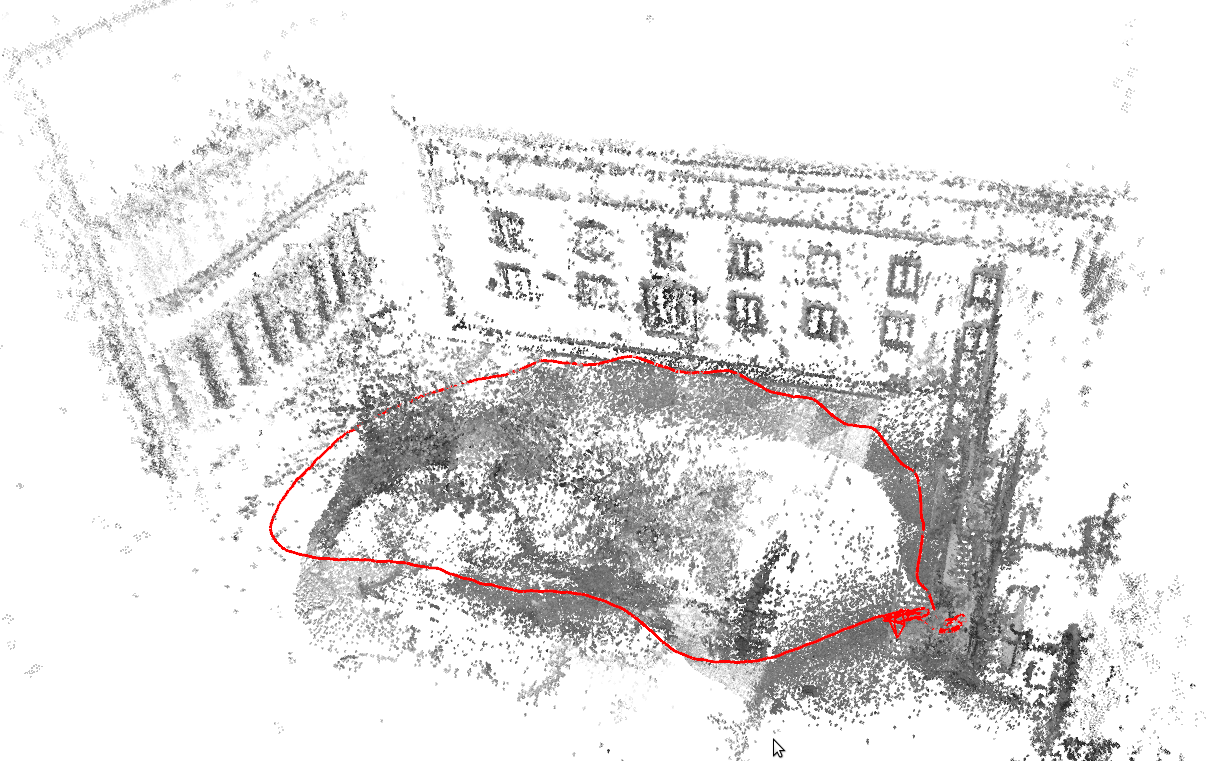}

        \end{center}
\end{figure}
We extend DSO to make seamless, full use of fisheye images (Fig.~\ref{top}) and to jointly optimize for the model parameters including camera intrinsics and extrinsics, sparse point depths and affine brightness parameters.

In experiments, we evaluate our approach on a benchmark of image sequences captured with a wide FoV fish-eye lens.
We compare our approach to other state-of-the-art VO approaches such as DSO or LSD-SLAM~\cite{LSD} and validate that our method outperforms the previous state-of-the-art on benchmark datasets.

The paper is organized as follows: We first review the state-of-the-art in Sec.~\ref{sec:relatedwork}. 
In Secs.~\ref{sec:notation} and~\ref{sec:cameramodels}, we introduce notation, the pinhole and the unified omnidirectional camera model. 
In Sec.~\ref{sec:overview}, we describe the pipeline of our omnidirectional DSO method. Our method is based on Direct Sparse Odometry~\cite{DSO} and integrates the unified omnidirectional camera model similar to~\cite{Omni}. We give a brief review of DSO and continue by detailing distance estimation along the epipolar curve with the unified omnidirectional camera model. 
In Sec.~\ref{sec:experiments}, we evaluate the performance of our method on publicly available datasets and compare it to the state-of-the-art.

\section{RELATED WORK}
\label{sec:relatedwork}



{\bf Indirect visual-odometry methods:}
Early works on visual odometry and visual simultaneous localization and mapping (SLAM) have been proposed around the year 2000 \cite{nister2004visual,civera2008inverse,jin_et_al2000} and relied on matching interest points between images to estimate the motion of the camera. While visual odometry methods focus on incremental real-time tracking of the camera pose with local consistency, SLAM approaches jointly estimate a globally consistent trajectory and map.
Many of these approaches were based on probabilistic filtering (e.g.~\cite{monoslam,chiuso2002_sfmcausally}).
For example, MonoSLAM proposed by Davison~\emph{et al.}~\cite{monoslam} is a real-time capable approach based on the Extended Kalman Filter.
However, since the landmark estimates are part of the filtered state space, the method is only capable to map small work-spaces due to computational limitations.
A technical breakthrough occurred in 2007 when Klein \emph{et al.} proposed PTAM~\cite{klein2007parallel}, a keyframe-based approach that performs tracking and mapping in separate threads. 
Similarly, many current VO/SLAM algorithms also use keyframes and apply multithreading to perform locally consistent tracking and mapping in real-time while optimizing for global consistency in a slower SLAM optimization layer.
ORB-SLAM~\cite{ORB} is the current state-of-the-art indirect and keyframe-based visual SLAM algorithm, which performs full bundle adjustment in a separate optimization layer.

{\bf Direct visual-odometry methods:}
More recently, direct methods have gained popularity for VO and SLAM. 
Direct methods avoid the extraction of geometric features such as keypoints but directly estimate odometry and 3D reconstruction from pixel intensities. 
Since they do not compress the image content to a small set of typically hand-crafted features, direct methods can use much more information in an image such as edges or shaded surfaces. 
This enables more dense 3D reconstructions while indirect methods only produce sparse point reconstructions. 
Direct visual odometry methods have also been proposed for RGB-D cameras, e.g.~\cite{kerl}.
The method extracts RGB-D keyframes and tracks the camera motion towards the recent keyframe using direct image alignment based on the measured depth.
LSD-SLAM~\cite{LSD} has been the first direct visual SLAM approach for monocular cameras that is capable of mapping large scale environments in realtime. 
It tracks the camera motion, produces a semi-dense map and performs pose graph optimization to obtain a consistent global map. 
The semi-dense maps can be adapted to a variety of uses such as surface estimation in AR, 3D object recognition and semantic labeling~\cite{schops2014semi,semantic,cnn}.

In pose graph optimization, the individual direct image alignment measurements are aggregated in a relative pose measurement between keyframes.
This neglects the fine-grained correlations of the direct measurements and requires linearization and Gaussian approximations to condense the measurement.
Recently, Direct Sparse Odometry (DSO) has been proposed by Engel \emph{et al.} \cite{DSO}.
In contrast to LSD-SLAM, DSO jointly optimizes multiple model parameters such as camera intrinsics/extrinsics, affine brightness parameters and depth in realtime within a window of keyframes by using a sparse set of points in each keyframe. 
This approach currently defines the state-of-the-art performance among visual odometry methods in terms of trajectory accuracy.

{\bf Limitation of monocular visual-odometry:}
Since monocular visual odometry estimates camera motion and scene reconstruction with a single camera, scale is invariably ambiguous and prone to drift. To recover metric scale, VO/SLAM methods are typically extended with additional sensors such as stereo camera, depth sensors or IMUs~\cite{ORB2,kerl,schur}. More recently, CNN-based depth predictions are combined with monocular visual SLAM~\cite{cnn}. In DSO and in our methods, due to the windowed optimization and marginalization, scale drift is comparably smaller than in tracking-based VO such as the VO front-end in LSD-SLAM.

{\bf Visual-odometry methods with omnidirectional camera models:}
To benefit from a larger FoV, VO and SLAM methods have also been extended for wide-FoV cameras~\cite{scaramuzza2008_omnivo,gutierrez2011_omnivis,Multicol,silpa2005visual}. 
In particular, Omnidirectional LSD-SLAM~\cite{Omni} has been the first direct visual SLAM approach for fisheye cameras which runs in real-time.
By incorporating the unified omnidirectional camera model, it works even for cameras with a FoV of more than 180 degree. 
In our approach, we also use the unified omnidirectional camera model, but optimize for a multitude of parameters such as camera intrinsics/extrinsics, affine brightness parameters and depth within a window of frames in DSO.
We demonstrate how the combination of an optimization window with the extended FoV improves performance over direct baseline methods such as DSO and LSD-SLAM.
Zhang~\emph{et al.}~\cite{survey} developed  Semi Direct Visual Odometry (SVO)~\cite{forster2014svo} for fisheye cameras and compared the performance with different FoV under the same image resolution. 
According to their paper, the optimal FoV also depends on the environment so that a wider FoV does not always improve results.
In indoor environments, however, they found that the wider FoV tends to increase performance.

{\bf Contribution:}
In this paper we present an omnidirectional extension of Direct Sparse Odometry. This is the first fisheye-based direct visual odometry which runs in real time and jointly optimizes multiple model parameters - camera pose, depth of points, camera intrinsics and affine brightness parameters.

\section{NOTATION}
\label{sec:notation}
We basically follow the notation in~\cite{Omni}:
We denote scalars $u$ with light lower-case letters, while light upper-case letters represent functions $I$. 
For matrices and vectors, we use bold capital letters $\bf R$ and bold lower-case letters $\bf x$, respectively. 
With $\mathbf{u} = \left[ u, v \right]^T \in \Omega \subset \mathbb{R}^2$ we will generally denote pixel coordinates, where $\Omega$ denotes the image domain. 
Point coordinates in 3D are denoted as $\mathbf{x} = \left[ x, y, z \right]^T \subset \mathbb{R}^3$. The operator $\left[\;\right]_i$ extracts the~$i$-th row of a matrix or vector.
We represent camera poses by matrices of the special Euclidean
group~${\bf T}_i \subset SE(3)$.
They transform a 3D coordinate from
the camera coordinate system to the world coordinate system.
In general, a camera projection function is a mapping $\pi : \mathbb{R}^3 \to \Omega$. 
Its inverse $\pi^{-1} : \Omega \times \mathbb{R}^+ \to \mathbb{R}^3$ unprojects image coordinates using their inverse distance $d \in  \mathbb{R}^+$. 
Camera frames are centered at~$C$ and the optical axis is along the z-axis pointing forward in positive direction.
For optimization, we represent a 3D point by its image coordinate~$\mathbf{p}$ and inverse distance~$d_p$ in its host keyframe in which it is estimated.

\section{CAMERA MODELS}
\label{sec:cameramodels}

In the following, we describe the two camera models used in this paper; the pinhole model and the unified omnidirectional model.

\subsection{Pinhole Model}
\begin{figure}
\begin{center}
		\includegraphics[width=7.5cm]{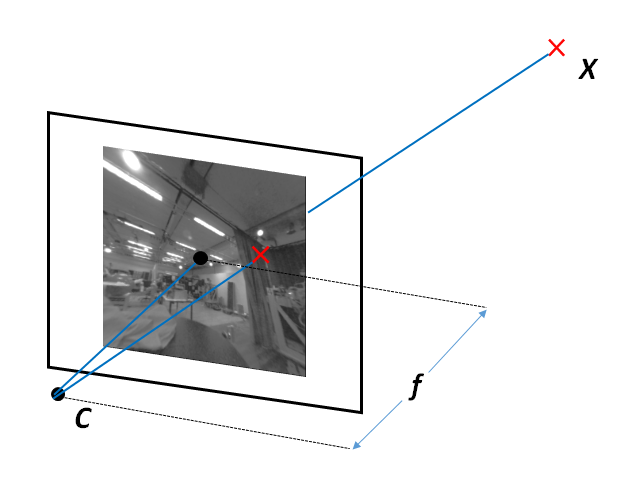}
		\caption{Pinhole camera model. The image coordinate of a 3D point is found through direct projection onto the image plane through the camera center $C$.}
		\label{fig:pinhole}
\end{center}
\end{figure}
The pinhole camera model is the most popular camera model in literature.
Each 3D point is projected onto a normalized image plane located at~$z = 1$ and then linearly transformed into pixel coordinates. 
This is mathematically formulated as
\begin{eqnarray}
\pi_u(\bf{x})=
\begin{bmatrix}
f_x & 0\\ 
0 & f_y\\
\end{bmatrix}
\begin{bmatrix}
    x/z \\ 
    y/z\\
\end{bmatrix}
+
\begin{bmatrix}
c_x \\ 
c_y\\
\end{bmatrix},
\end{eqnarray}
where $f_x, f_y$ are the focal lengths, and $c_x, c_y$ is the principal point. 
The projection model is illustrated in Fig.~\ref{fig:pinhole}.

This is the most simple model because the projection
function is linear in homogeneous coordinates. However, this does not consider the nonlinear image projection of fisheye images and is not suitable for wide FoV cameras. 
Radial and tangential distortion functions can be applied to remove small non-linear distortions, however, the pinhole projection assumes that the measured 3D points are beyond the image plane, i.e. their depth is larger than the focal length. This limits the
field-of-view below 180 degree.

\subsection{Unified Omnidirectional Model}
We use the unified omnidirectional camera model which has been originally proposed in \cite{Omnimodel} for a wide FoV fish-eye camera. The major advantages of this model are; (1) it can accurately model the geometric image formation for a wide range of imaging devices and lenses, (2) the unprojection function ${\pi}^{-1}$ can be expressed in closed-form.
\begin{figure}
		\includegraphics[width=5.3cm,bb=0 0 250 250]{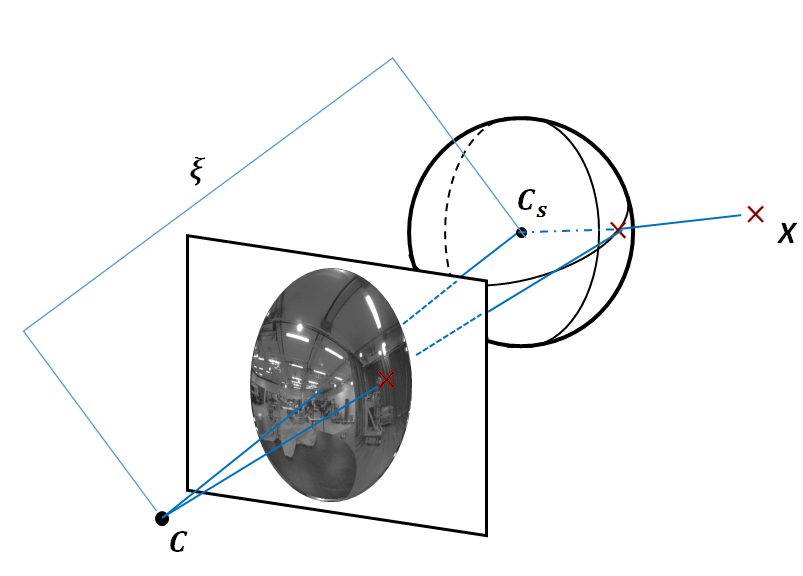}
		\caption{Unified Omnidirectional Camera Model. The image coorindate of a 3D point is found by first projecting it on the unit sphere,
and then projecting it to image plane. The offset between camera center $C$ and unit sphere center $C_s$ is parameterized as $\xi$.}
		\label{omni}
\end{figure}
 A 3D point in Euclidean camera coordinates is first projected onto a camera-centered unit sphere (see Fig.~\ref{omni}). Then the point is projected to an image plane as in the pinhole model through a center with an offset $-\xi$ along the $z$ axis. The model has five parameters, focal length $f_x, f_y$, camera centers $c_x, c_y$ and the distance between camera center and unit sphere center $\xi$. 

\ \ The projection of a point is computed as 
\begin{eqnarray}
    \pi_u(\bf{x})=
    \begin{bmatrix}
    f_x{\frac{x}{z+||\bf x||\xi}}\\[7pt]
    f_y{\frac{y}{z+||\bf x||\xi}}\\
    \end{bmatrix}
    +
    \begin{bmatrix}
    c_x \\ 
    c_y\\
    \end{bmatrix}
\end{eqnarray}

where $||\bf x||$ denotes the Euclidean norm of $\bf x$. The unprojection function for this model is
\begin{eqnarray}
\begin{split}
\pi_u^{-1}(&{\bf{u}},d)\\&=\frac{1}{d}\left(
\frac{\xi+\sqrt{1+(1-\xi^2)(\tilde{u}^2+\tilde{v}^2)}}{\tilde{u}^2+\tilde{v}^2+1}
\begin{bmatrix}
\tilde{u}\\ 
\tilde{v}\\
1\\
\end{bmatrix}
-
\begin{bmatrix}
0 \\ 
0 \\
\xi \\
\end{bmatrix}
\right)
\end{split}
\end{eqnarray}
where
\begin{eqnarray}
\begin{bmatrix}
\tilde{u}\\ 
\tilde{v}\\
\end{bmatrix}
=
\begin{bmatrix}
(u-c_x)/f_x \\ 
(v-c_y)/f_y\\
\end{bmatrix}.
\end{eqnarray}
Note that for $\xi=0$ the model reduces to the pinhole model. We combine the unified omnidirectional model with a small radial-tangential distortion model to correct for lens imperfections. The model is used to undistort the raw images before applying the unified omnidirectional model.

\section{System Overview}
\label{sec:overview}
\begin{figure}[h]
     \begin{center}
		\includegraphics[width=8.5cm]{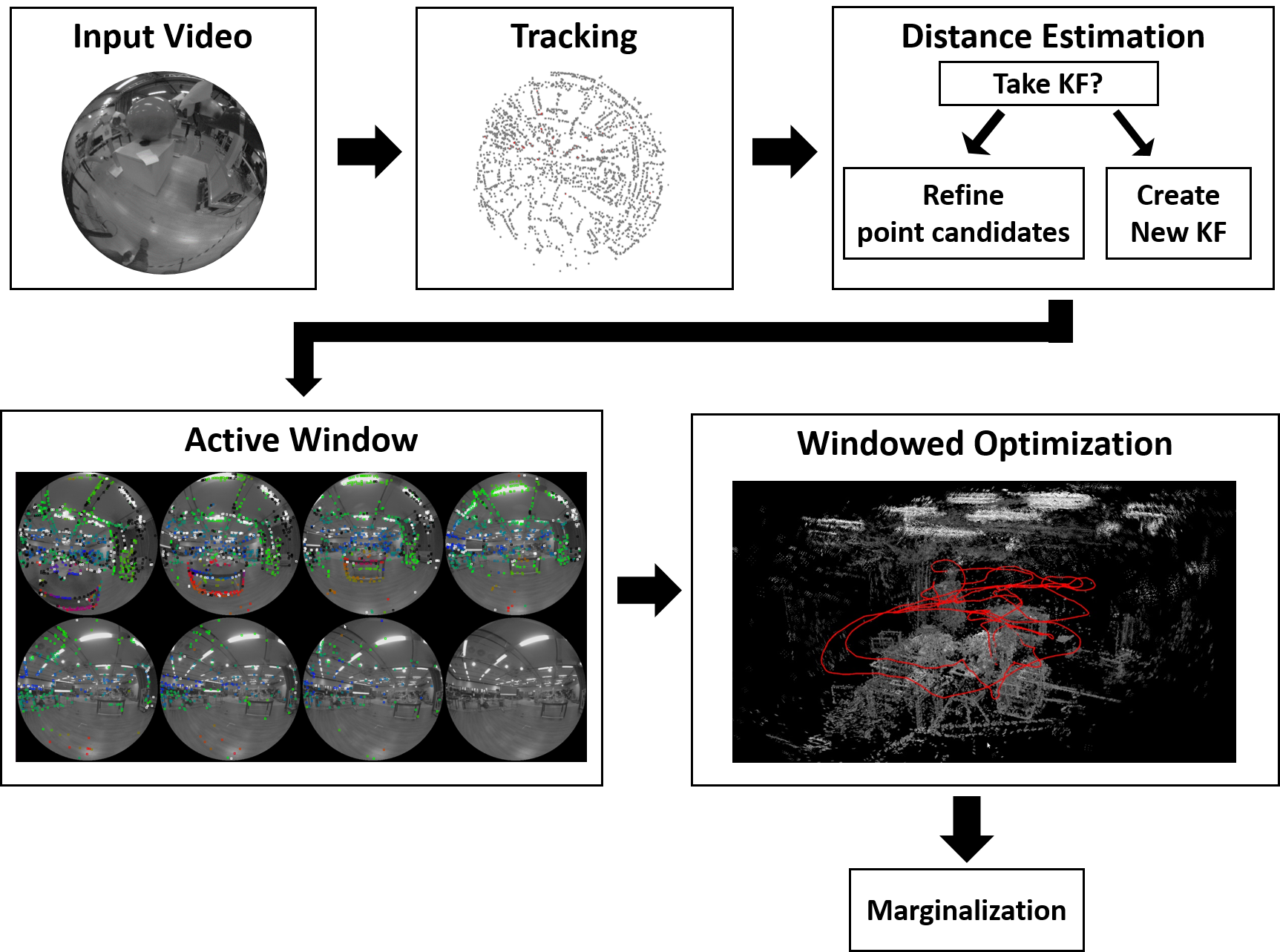}
				\caption{System overview of omnidirectional DSO (OmniDSO).}
		\label{overview}
		        \end{center}
\end{figure}
\subsection{Model Formulation}
DSO jointly optimizes camera poses, point depths and affine brightness parameters in a window of recent frames. As a direct method it optimizes for photometric consistency. DSO also takes the photometric calibration of the image formation process into account. 
\par The energy function which represents the photometric error between two frames is formulated as
\begin{eqnarray}
E_{{\bf{p}}j} := \sum_{{\bf{p}}\in{N_p}}w_p\|\:(I_j[{\bf{p'}}]-b_j)-\frac{t_je^{a_j}}{t_ie^{a_i}}(I_i[{\bf{p}}]-b_i)\:\|_{\gamma}
\end{eqnarray}
where we measure the photometric error of a point $\mathbf{p} \subset \Omega_i$ in reference frame $I_i$ with respect to a target frame $I_j$ through the weighted sum of squared differences (SSD) over a small pixel neighborhood $N_p$. $w_p$ is a gradient dependent weighting. $t_i$, $t_j$ are the exposure times of the images $I_i$, $I_j$; $a$ and $b$ are affine brightness correction factors; and $||\cdot||_\gamma$ denotes the Huber norm. ${\bf{p'}}$ is the reprojected point position of ${\bf{p}}$ with inverse distance $d_p$. ${\bf{p'}}$ is given by
\begin{eqnarray}
 {\bf{p'}}=\pi({\bf{R}}\pi^{-1}({\bf{p}},d_p)+{\bf{t}})
\end{eqnarray}
with
\begin{eqnarray}
\begin{bmatrix}
{\bf{R}} &{ \bf{t}}\\ 
0 & 1\\
\end{bmatrix}
:={\bf{T_j}}\:{\bf{T_i}}^{-1}.  
\end{eqnarray} 
The photometric error terms of the active window of frames are
\begin{eqnarray}
\label{energy}
E_{photo} := \sum_{i\in{F}}\;\sum_{{\bf{p}}\in{P_i}}\sum_{j\in{obs({\bf{p}})}} E_{{\bf{p}}j}
\end{eqnarray}
where $F$ is the set of frames in the active window, $P_i$ are the points in frame $i$, and $obs({\bf{p}})$ is the set of frames which observe the point ${\bf{p}}$.
For tracking, this error function is minimized with respect to the relative camera pose ${\bf{T_{ij}}}$ between ${\bf{T_i}}$ and ${\bf{T_j}}$. For window optimization, the function is optimized for all variables $({\bf{T_i}},{\bf{T_j}},d,{\bf{c}},a_i,a_j,b_i,b_j)$, where ${\bf{c}}$ are camera intrinsic parameters. Different to~\cite{DSO}, we parametrize points with the inverse
distance $d = |\bf {x}|^{-1}$ instead of inverse depth. This allows us to model
points behind the camera as well (s. Fig.~\ref{overview} for an overview). 

\subsection{Distance Estimation along the Epipolar Curve}\label{sec:episearch}
\begin{figure}[thpb]
     \begin{center}
		\includegraphics[width=8.5cm]{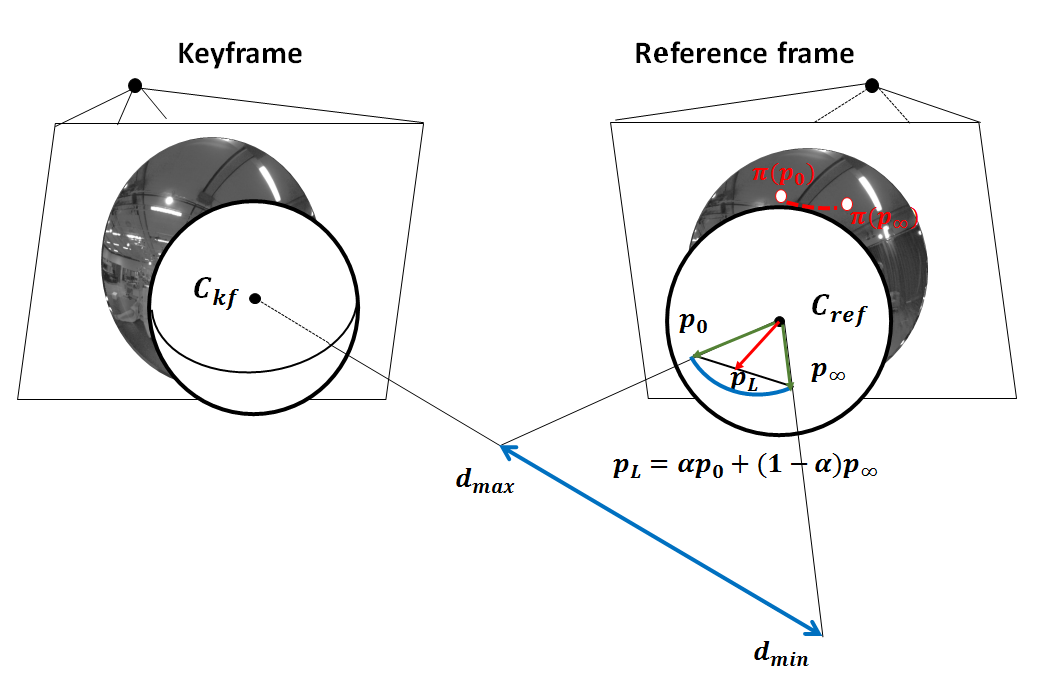}
				\caption{Stereo Matching with the unified omnidirectional camera model: Correspondence search for stereo matching is performed by mapping the distance search interval along a ray to a directional interval on the projection unit sphere of the other camera.}
		\label{DSO}
		        \end{center}
\end{figure}
Once a frame is successfully tracked, we perform stereo matching to refine the inverse distance of candidate points.
When a candidate point gets included into the photometric bundle adjustment, this estimated distance serves as an initialization.
DSO searches for corresponding points along the epipolar line similar to \cite{Semidense}. However, when performing stereo matching on fisheye images using the unified omnidirectional model, rays through camera center and pixels project no longer to epipolar lines but curves (more precisely they are conics \cite{Omnimodel}).

We now describe the mathematical formulation of the epipolar curve.
Similar as in~\cite{Omni}, we define two points $\bf p_0,\; p_\infty\in\mathbb{R}^3$ which lie on the 
unit sphere around a projective center $C_{ref}$ and correspond to the maximum and minimum inverse distance $d_{max}, d_{min}$ of the search interval,
\begin{eqnarray}
{\bf{p_0}} := \pi_s({\bf{R}}\pi_u^{-1}({\bf{p}},d_{min})+{\bf{t}})\\
{\bf{p_\infty}} := \pi_s({\bf{R}}\pi_u^{-1}({\bf{p}},d_{max})+{\bf{t}}).
\end{eqnarray}

Here, the function $\pi_s$ projects the 3D points onto the unit sphere. $\pi^{-1}_u$ is the unprojection function of the unified model, and $\bf{p}$ is the
pixel in the keyframe we are trying to match. We then express the linear
interpolation of these points with $\alpha \in [0, 1]$ as
\begin{eqnarray}
{{\bf{p}}_L(\alpha)} := \alpha{\bf{p}}_0+(1-\alpha){\bf{p}}_\infty
\end{eqnarray}
We find the epipolar curve by projecting this line to the target image,
\begin{eqnarray}
{\bf u}_L(\alpha) :=  \pi_u({\bf{p}}_L(\alpha))
\end{eqnarray}
We then search correspondences along the epipolar curve by starting at ${\bf u}_L(0)$ and incrementing $\alpha$. The increment in $\alpha$ for 1 pixel in the image is determined by first-order Taylor approximation of ${\bf u}_L$ as
\begin{eqnarray}
{{\delta\alpha}} :=  \|\;{\bf{J}}_{{\bf{u}}_L}|_\alpha\;\|^{-1}
\end{eqnarray}
This value needs to be re-calculated for each increment while for the pinhole camera model a constant step size can be used for epipolar line search. However, in DSO and LSD-SLAM, a distance prior is available from previous frame estimates or through initialization. Hence, the search interval is typically small and real-time computation is facilitated.


\subsection{Frame Management}
\label{sec:framemanagement}

DSO maintains a constant number of $N_f$ active keyframes in the optimization window (e.g. $N_f = 7$). 
It keeps track of the camera motion in every new frame by tracking it towards the latest keyframe and its sparse distance map (step 1). 
If the changes in the observed scene are too large towards the latest keyframe, a keyframe is created from the new frame (step 2).
Afterwards, we marginalize one or more frames to keep the number of keyframes constrained (step 3).

\subsubsection{Initial Frame Tracking}
For tracking, conventional direct image alignment is performed in a 5 level image pyramid. The scene and brightness change is continuously estimated and if the change is bigger than a certain threshold value, the frame is selected as a new keyframe. 
\subsubsection{Keyframe Creation} 
\par When a keyframe is created, candidate points are selected considering space distribution and image gradient. 
We initialize the inverse distance estimate of these candidate points with a large variance that corresponds to a range from zero to infinity. After each subsequent new frame has been tracked towards this new key frame, the inverse distance estimates are refined using observations in the new frame which we obtain through the epipolar search (Sec.~\ref{sec:episearch}). 
\subsubsection{Keyframe Marginalization}
\par When the number of active keyframes grows above $N_f$, old points and frames are removed from the active window considering the number of visible points and frame distribution.
Let us denote the active keyframes with $I_1,\ldots,I_n$, while $I_1$
is the newest and $I_n$ the oldest keyframe.
Our marginalization strategy follows~\cite{DSO}: 
a. We never marginalize the latest two keyframes ($I_1$, $I_2$).
b. We marginalize a frames if the percentage of its points that are visible in $I_1$ drops below 5\%.
c. If the number of active keyframes grows above $N_f$, we marginalize 
one keyframe based on a heuristic distance score~\cite{DSO}. The score favors active keyframes that are spatially distributed close to the latest keyframe.
Finally, candidate points are activated and added to the optimization.

\subsection{Windowed Optimization}
Our windowed optimization and marginalization policy follows \cite{DSO}. As we formulated in (\ref{energy}), joint optimization is done for all activated points over all active keyframes.
Nonlinear optimization for the photometric error function is performed using the Gauss-Newton algorithm \cite{DSO}. 
All the variables including camera pose, inverse distance of active points, camera intrinsic parameters, and affine brightness parameters are jointly optimized.
After the minimization of the photometric error, we marginalize old keyframes and points using
the Schur complement~(\cite{DSO,schur}) if the number of active keyframes in the optimization window grows beyond $N_f$. 
We keep the optimization problem sparse by first marginalizing those points that are unobserved in the two latest keyframes. We also marginalize the points which are hosted in the keyframe which will be marginalized.
Afterwards, the keyframe
is marginalized and removed from the optimization window. 

\begin{figure*}[h]

        \begin{center}
		\includegraphics[width=16.5cm]{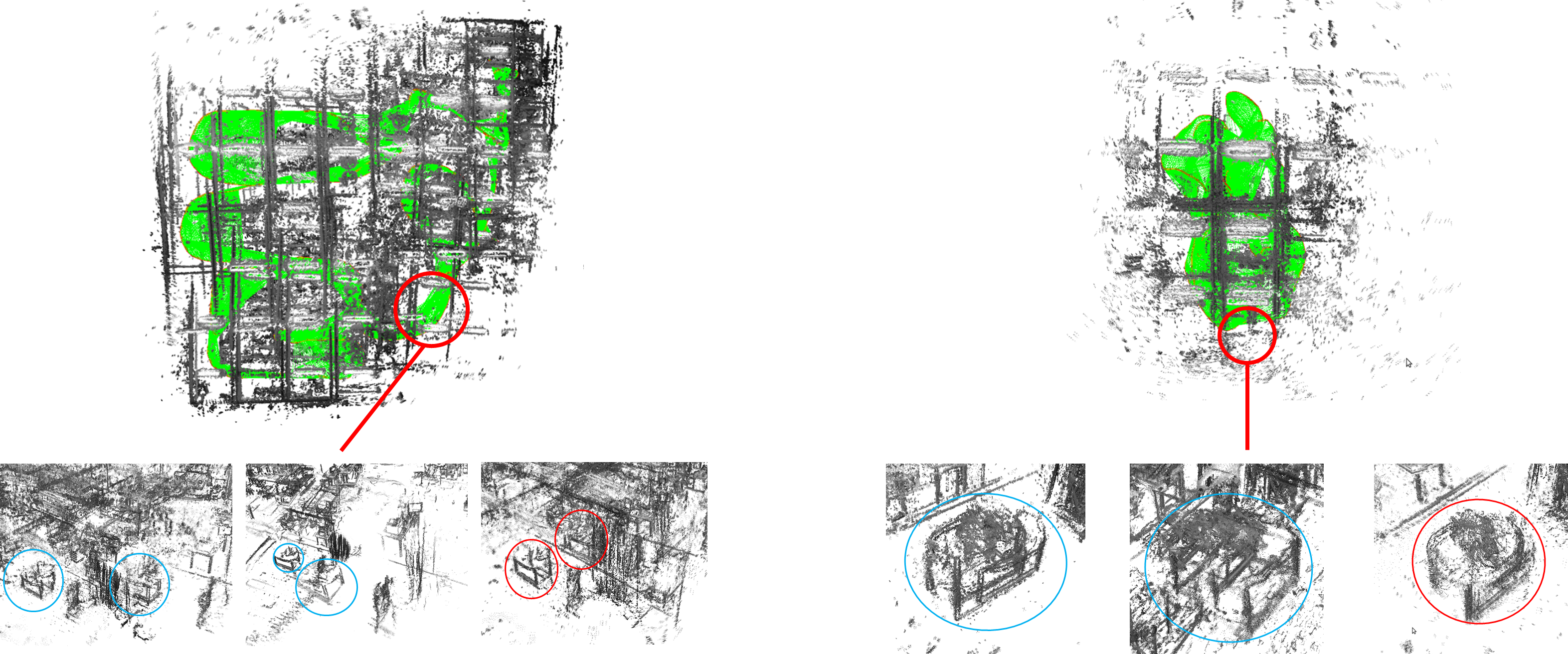}
				\caption{Accumulated drift (left: normal DSO, middle: OmniLSD (VO), right: OmniDSO). Compared to the other 2 methods, OmniDSO demonstrates better performance both in terms of translation and scale drift.}
		\label{pointcloud}
        \end{center}
\end{figure*}


\section{Evaluation}
\label{sec:experiments}
We provide an experimental evaluation of our method both in terms of accuracy and robustness. We perform a quantitative comparison against the state of the art visual odometry methods on public benchmark
datasets. We also qualitatively assess the benefit of a wider FoV.
We used two public datasets for evaluation: TUM SLAM for Omnidirectional Cameras dataset first employed in~\cite{Omni} as a small-scale indoor benchmark and Oxford RobotCar dataset \cite{RobotCarDatasetIJRR} as a large-scale outdoor benchmark.

\subsection{TUM SLAM for Omnidirectional Cameras Dataset}
The TUM omnidirectional dataset provides wide FoV fisheye images of indoor scenes. It also includes ground truth trajectory data recorded with a motion capture system and calibrated unified omnidirectional model camera parameters. The dataset consists of 5 indoor sequences with rapid and handheld motion. The camera is global shutter with a 185$^\circ$ FoV fisheye lens and recorded images of resolution $1280\times1024$.
We cropped and scaled the images to a $480\times480$ resolution centered around the principal point.
With this dataset, we compared 5 direct visual odometry algorithms; normal DSO~\cite{DSO}, omnidirectional DSO (our method), normal LSD-SLAM~\cite{LSD} (without loop-closing), omnidirectional LSD-SLAM~\cite{Omni} (without loop-closing), and semi-direct visual odometry (SVO ~\cite{forster2014svo}). Note that we turned off loop-closing of LSD-SLAM to evaluate the performance of its underlying visual odometry in terms of the overall drift per sequence.

\subsubsection{Accuracy Comparison}
\begin{figure}[h]
     \begin{center}
		\includegraphics[width=8.5cm]{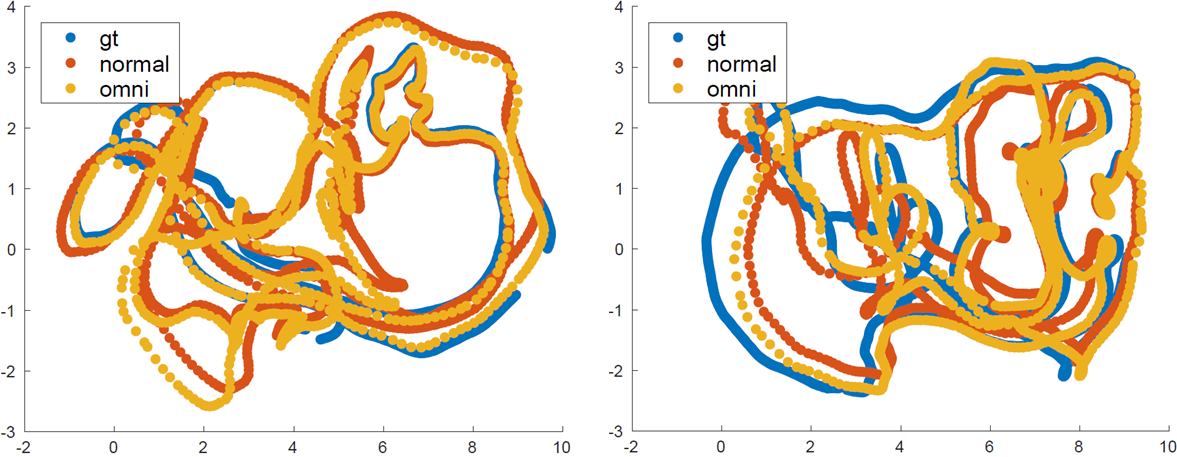}
				\caption{{\bf{Top: }}Horizontal trajectory estimates for T1 and T3. 
}
		\label{traj}

        \end{center}
\end{figure}
Following the evaluation methodology in \cite{Omni} we measured
the translational root mean square error (RMSE) between the estimated and the ground-truth camera translation for each evaluated sequence.
The estimated camera position is calculated for all keyframes, and Sim(3) alignment with the ground-truth trajectory data is performed. 
Since the multi-threaded execution of DSO introduces non-deterministic behavior, we ran the algorithm 5 times for each sequence per method, then took the median RMSE. The results are shown in Table~\ref{table}. Some representative visual results are shown in Fig.~\ref{pointcloud} and \ref{traj}.
Table~\ref{table2} shows the length of the trajectory estimated by OmniDSO.

\par  We make two observations from Table~\ref{table}: First, DSO is more robust and accurate than SVO and LSD-SLAM without loop closure. Since the dataset scene contains a lot of small loops, this contributes to SLAM performance. However, as a pure visual odometry, DSO shows much better performance. This means sparse joint bundle adjustment and windowed optimization increases the performance of direct visual odometry. Second, the use of the unified omnidirectional camera model further improves the performance both for DSO and LSD-SLAM. Although for some sequences a clear performance improvement cannot be observed, considering the average result over all sequences, fisheye visual odometry demonstrates a clear advantage over using the pinhole camera model (approx. 0.157\,m improvement of OmniDSO over DSO). Our OmniDSO incorporates both of these benefits and outperforms other existing visual odometry methods. From Tables~\ref{table} and~\ref{table2}, we can clearly see the performance difference in T5, which has the longest trajectory among sequences.

\begin{table}[h]
\caption{Absolute RMSE in meters. DSO outperforms other methods and especially our OmniDSO shows the smallest RMSE in average.}
\label{table}
\begin{center}
\scalebox{0.9}[1.0]{
\begin{tabular}{|c|c|c|c|c|c|c|c|}
\hline
 & DSO & \bf{Ours} & LSD & OmniLSD &SVO &OmniLSD-SLAM\\
\hline\hline
T1 & 0.243 & \bf0.144 & 0.751 & 0.873 & 1.22 &(0.053)\\
\hline
T2 & \bf0.450 & 0.497 & 1.43 & 1.22 & 0.980  &(0.051)\\
\hline
T3 & 0.499 & \bf0.258 & 1.43 & 0.551 & 1.28  &(0.046)\\
\hline
T4 & \bf0.240 & 0.254 & 0.731 & 0.752 & 0.734  &(0.045)\\
\hline
T5 & 1.47 & \bf0.960 & 1.91 & 1.69 & 3.06  &(0.036)\\
\hline
\hline
Avg. & 0.580 & \bf0.423 & 1.25 & 1.02 & 1.46 &(0.046)\\
\hline
\end{tabular}
}
\end{center}

\caption{Generated trajectory length in meters.}
\label{table2}
\begin{center}
\scalebox{1.0}[0.9]{
\begin{tabular}{|c|c|c|c|c|}
\hline
T1 & T2 & T3 & T4 &T5\\
\hline
101 & 83.5 & 87.5 & 93.5 & 138 \\
\hline
\end{tabular}
}
\end{center}
\end{table}

\subsubsection{Benefit of Large Field of View}
\begin{figure}[h]
     \begin{center}
		\includegraphics[width=8.5cm]{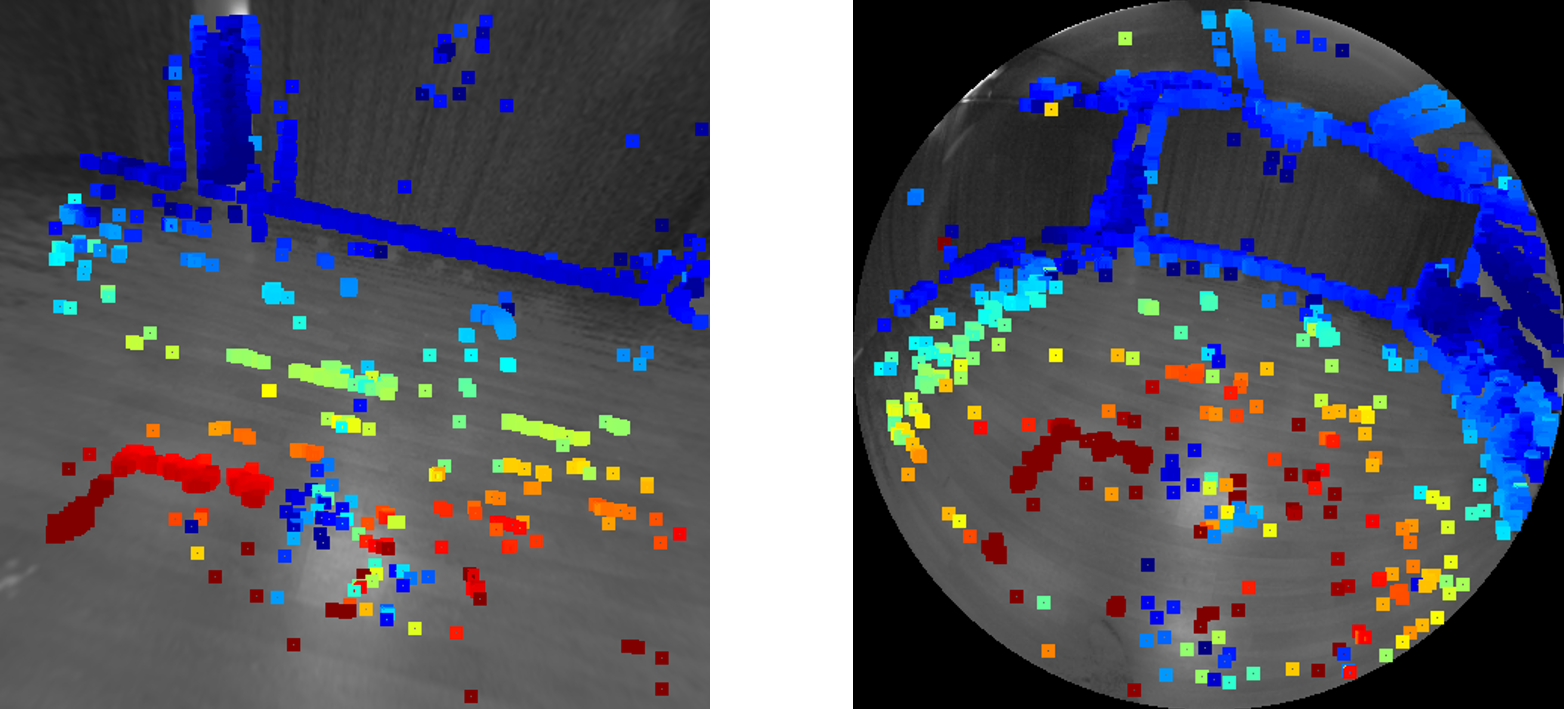}
				\caption{Active points for optimization in a textureless scene  (left: DSO, right: OmniDSO). With a wider field of view, it is less likely that the observed scene exhibits a degeneracy in any spatial direction. The left image exhibits an intensity constancy along the diagonal and hence motion estimation in that direction is not constrained.}
		\label{texture}
		        \end{center}
     \begin{center}
		\includegraphics[width=8.5cm]{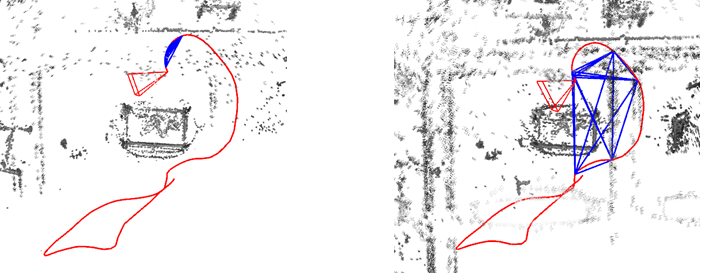}
				\caption{Active keyframes in optimization window after camera rotation  (left: DSO, right: OmniDSO). In our method, keyframes are not marginalized even after big camera rotation.}
		\label{rot}
		        \end{center}
         \begin{center}
		\includegraphics[width=8.0cm]{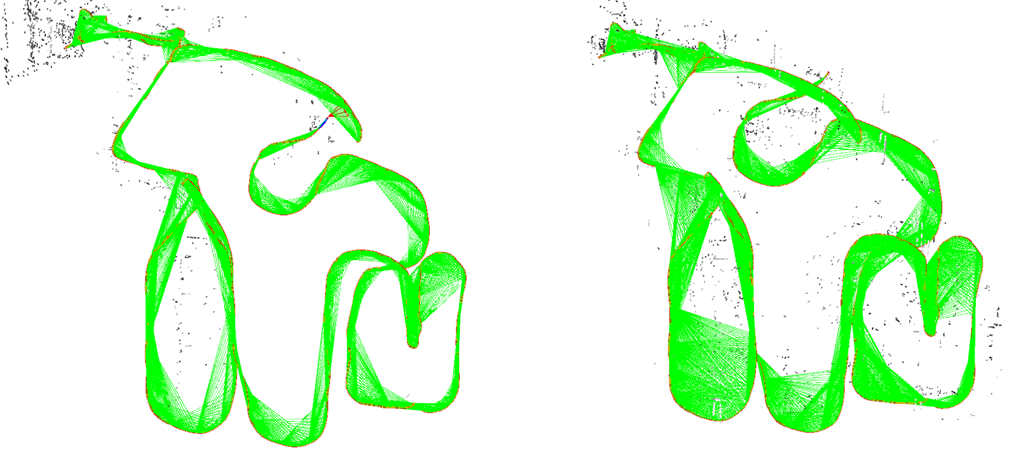}
				\caption{Constraints between keyframes (green lines) during optimization (left: DSO, right: OmniDSO). OmniDSO maintains more spatially distributed keyframes within the optimization window.}
		\label{marg}
		        \end{center}
\end{figure} 
One of the major advantages of using wide FoV is that the image is more likely to contain strong gradient pixels which are beneficial for visual odometry. Fig.~\ref{texture} shows active points in a keyframe for the textureless scene in the T5 sequence for DSO with the pinhole and the unified omnidirectional camera model. 
Note that DSO uses points with strong image gradient.
When comparing the two camera models, it is apparent that the omnidirectional camera model can track on a larger spatial distribution of points with a larger variety of gradient directions, while the pinhole camera model only observes a smaller inner part of the images.

Another benefit of using the omnidirectional model is the bigger image overlap between frames. 
Fig.~\ref{rot} and \ref{marg} show the estimated trajectory and constraints between active keyframes in the current optimization window in the same scene for DSO and OmniDSO. 
As we described in \ref{sec:framemanagement}, a keyframe is marginalized from the window when the number of visible points from the current frame falls below 5\%. 
Due to the increased image overlap, the omnidirectional camera model allows for maintaining a keyframe longer in the optimization window, thus increasing the spatial distribution of keyframes and the effective part of the scene that is observed within the window compared to DSO with the pinhole model.
To evaluate this overlap effect numerically, we tested different maximum numbers of keyframes for the windowed optimization and compared the results. The tested number of keyframes ($N_f$) were $N_f=7$ (default), $N_f=5$ and $N_f=3$. Table~\ref{keyframes} shows the result of reducing keyframes and Table~\ref{comparison} shows the performance difference. 
With Kf5-Kf7 we denote the difference in RMSE between $N_f=5$ and $N_f=7$.
The smaller the number the more robust the approach is to keyframe number reduction. 
Table~\ref{comparison} shows that the performance declines as the keyframe number decreases for both methods. 
However, as shown in Table~\ref{comparison}, the decrease of our method (Omni DSO) is smaller than that of normal DSO. 
This demonstrates that the bigger visual overlap due to the use of the omnidirectional model contributes to maintaining performance even if less keyframes are used.

\begin{table}[h]
\centering
\caption{Absolute RMSE for different number of keyframes in meters. For all keyframe number settings, OmniDSO shows smaller average RMSE.}
\label{keyframes}
\scalebox{1.0}[1.0]{
\begin{tabular}{c|c|c|c|c|c|c|}
\cline{2-7}
                         & \multicolumn{3}{c|}{Normal DSO} & \multicolumn{3}{c|}{Omni DSO} \\ \hline
\multicolumn{1}{|c|}{$N_f$}   & $7$      & $5$     & $3$     & $7$     & $5$     & $3$    \\ \hline
\multicolumn{1}{|l|}{T1} & 0.243     & 0.288    & 0.35     & 0.144    & 0.145    & 0.218   \\ \hline
\multicolumn{1}{|l|}{T2} & 0.450     & 0.451    & 0.658    & 0.497    & 0.569    & 0.779   \\ \hline
\multicolumn{1}{|l|}{T3} & 0.499     & 0.642    & 0.624    & 0.258    & 0.304    & 0.382   \\ \hline
\multicolumn{1}{|l|}{T4} & 0.240     & 0.223    & 0.672    & 0.254    & 0.246    & 0.416   \\ \hline
\multicolumn{1}{|l|}{T5} & 1.47      & 1.71     & 1.82     & 0.96     & 1.07     & 1.14    \\ \hline
\multicolumn{1}{|l|}{Avg.} & 0.580      &0.663     & 0.825     & \textbf{0.423}     & \textbf{0.467}     & \textbf{0.587}    \\ \hline
\end{tabular}
}

\end{table}

\begin{table}[h]

\centering
\caption{Performance difference between used number of keyframes in meters. The average performance decline of OmniDSO with respect to keyframe number change is smaller}
\label{comparison}
\scalebox{1.0}[1.0]{
\begin{tabular}{l|l|l|l|l|}
\cline{2-5}
                          & \multicolumn{2}{c|}{Kf5-Kf7} & \multicolumn{2}{c|}{Kf3-Kf7} \\ \hline
\multicolumn{1}{|l|}{}    & Normal   & Omni              & Normal   & Omni              \\ \hline
\multicolumn{1}{|l|}{T1}  & 0.045    & 0.001             & 0.107    & 0.074             \\ \hline
\multicolumn{1}{|l|}{T2}  & 0.001    & 0.072             & 0.208    & 0.282             \\ \hline
\multicolumn{1}{|l|}{T3}  & 0.143    & 0.046             & 0.125    & 0.124             \\ \hline
\multicolumn{1}{|l|}{T4}  & -0.017   & -0.008            & 0.432    & 0.162             \\ \hline
\multicolumn{1}{|l|}{T5}  & 0.240     & 0.110              & 0.35     & 0.18              \\ \hline
\multicolumn{1}{|l|}{Avg.} & 0.082   & \textbf{0.044}   & 0.244   & \textbf{0.164}   \\ \hline
\end{tabular}
}

\end{table}

\begin{table}[h]

\centering
\caption{Mean timing results (ms). As the keyframe number reduces, the computational cost becomes smaller.}
\label{timing}
\scalebox{1.0}[1.0]{
\begin{tabular}{c|c|c|c|c|c|c|}
\cline{2-7}
                         & \multicolumn{3}{c|}{Normal DSO} & \multicolumn{3}{c|}{Omni DSO} \\ \hline
\multicolumn{1}{|c|}{$N_f$}   & $7$      & $5$     & $3$     & $7$     & $5$     & $3$    \\ \hline
\multicolumn{1}{|l|}{Tracking} & 3.7     & 4.1    & 3.6    & 9.9    & 9.8    & 8.5  \\ \hline
\multicolumn{1}{|l|}{Mapping} & 53     & 43& 30    & 63    & 54   & 36   \\ \hline 
\end{tabular}
}
\end{table}

\subsubsection{Timing measurement}
Table~\ref{timing} shows the measured average time over the dataset (5 runs per sequence) in
milliseconds taken for tracking and mapping (windowed optimization) steps.
To measure these times, images have been processed at resolution $480\times480$.
We used a computer with Intel Core i7-7820HK CPU at 2.90\,GHz with 4 cores.
These results demonstrate the realtime capability of our method since each frame can be tracked at
at least 100\,Hz and mapped at more than 15\,Hz.
These results also show that the mapping process is sped up by reducing the number of keyframes. Even with less number of keyframes ($N_f=5$), Table \ref{keyframes} displays that our method still outperforms normal DSO in average with $N_f=7$ keyframes.  


\subsection{Oxford Robotcar Dataset}
\begin{figure}[h]
     \begin{center}
		\includegraphics[width=8.0cm]{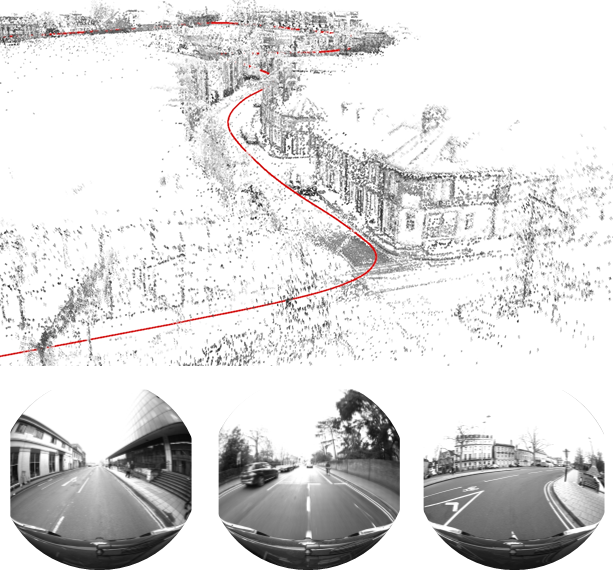}
				\caption{Example qualitative result of OmniDSO on the Oxford Robotcar dataset. The dataset contains a large-scale outdoor trajectory through Oxford.}
		\label{oxford_pc}
    \end{center}
\end{figure}

We also evaluated our algorithm on the Oxford Robotcar dataset in a large-scale outdoor scenario.
The dataset contains more than 100 repetitions of a consistent route at different times under different weather, illumination and traffic conditions. The images are collected from 6 cameras mounted on the vehicle, along with LIDAR, GPS and INS ground truth.
As an evaluation benchmark, we used videos taken by a rear-mounted global shutter camera with a 180$^\circ$ FoV fisheye lens. The raw image resolution is $1024\times1024$ and we cropped and scaled it to a $600\times600$ resolution.
We obtained camera intrinsic parameters using the Kalibr calibration toolbox \cite{furgale2013_kalibr} with the original checkerboard sequence.
The dataset has 3 types of routes (Full, Alternate and Partial). Full route covers the whole original route and consists of 5 sequences. Alternate route covers a different area and Partial route is a part of Full route. 
From this dataset, we selected sequences with overcast weather and less moving objects such as vehicles and pedestrians.
Full 1 is chosen from 2015/02/10, Full 2,3,5 are from 2014/12/09 and Full 4 is from 2015/03/17.
Alternate is from 2014/06/26 and Partial is from 2014/12/05. 
In the same way as for the indoor dataset, we measured the translational RMSE between the generated trajectory and the ground-truth data. We compared OmniDSO with DSO and monocular ORB-SLAM, the latter two using the pinhole camera model. We used the ORB-SLAM2 implementation (https://github.com/raulmur/ORB\_SLAM2).
Because the selected sequences do not contain loops, we can fairly compare VO and SLAM  without turning off the loop closure of SLAM. We similarly ran the algorithm 5 times for each sequence per method and took the median. The results and the trajectory length are shown in Table~\ref{table_oxford}. From the table, we observe that our method outperforms the other methods for all sequences. The performance difference tends to be more clear as the trajectory becomes longer.
Visual results are shown in Fig.~\ref{oxford_pc}.

\begin{table}[h]
\caption{Absolute RMSE and generated trajectory length in meters. Our OmniDSO shows the smallest RMSE.}
\label{table_oxford}
\begin{center}
\scalebox{0.90}[1.0]{
\begin{tabular}{|c|c|c|c|c|c|c|c|c|}
\hline
 & Full1 & Full2 & Full3 & Full4 &Full5 &Alternate &Partial\\
\hline\hline
ORB-SLAM & 12.1 & 34.2& 43.1 & 67.1 & 1.83 & 46.0& 89.5 \\
\hline
DSO & 10.0 & 26.4 & 27.4 &58.2 &0.987 &22.9 &60.1 \\
\hline
\bf{Ours} & \bf9.30 &\bf24.5 &\bf26.9 &\bf45.7 &\bf0.822 &\bf21.6 &\bf50.3 \\
\hline
\hline
Length &736 &1459 &1554	&1719 &204 &1003 &2433 \\
\hline
\end{tabular}
}
\end{center}
\end{table}

\section{CONCLUSIONS}

In this paper, we introduced real-time Direct Sparse Odometry for omnidirectional cameras. We first incorporate the unified omnidirectional camera model into DSO. Depth estimation is performed by efficiently searching along the epipolar curve incrementally. Camera pose, point depth and affine brightness parameters are jointly optimized to minimize photometric error within an active keyframe window. Then, we quantitatively evaluated the performance on 2 public benchmark datasets and demonstrated that our omnidirectional DSO yields better performance than other methods on the benchmark. 
We also qualitatively discussed the benefits of using a large field of view and quantitatively assessed the increase in robustness over using a pinhole camera model when reducing the number of keyframes in the DSO optimization window. 
Our omnidirectional DSO can make use of wide FoV fisheye camera images. Our combination of using a unified omnidirectional camera model and sparse windowed bundle-adjustment can outperform existing visual odometry methods. 
In future work, our method could be improved by adding global optimization and loop closure.

\addtolength{\textheight}{-12cm}   



\section*{ACKNOWLEDGMENT}

We thank Jakob Engel and Keisuke Tateno for their advice and fruitful discussions.


{\small
\bibliographystyle{IEEEtran}
\bibliography{abbrev_short,egbib}
}

\end{document}